\documentclass[letterpaper, 10 pt, conference]{ieeeconf}  % Comment this line out if you need a4paper

\IEEEoverridecommandlockouts                              % This command is only needed if 
\usepackage{eso-pic}
\usepackage{amsmath, amssymb, amsfonts}

\usepackage{subcaption}
\usepackage{xcolor}
\usepackage{xspace}
\usepackage{graphicx}
\usepackage{placeins}
\usepackage{paralist}
\usepackage{tikz}
\usetikzlibrary{automata, positioning, arrows, calc, fit, shapes}
\tikzset{initial text={}} % sets the text that appears on the start arrow

\usepackage{amsthm}

\usepackage{hyperref}  %hyperref still needs to be put at the end!
\usepackage{cleveref}
\newtheorem{problem}{Problem}
\newtheorem{myexam}{Example}

\newtheorem{definition}{Definition}

\newcommand{\events}{\Pi}
\newcommand{\event}{e}

\newcommand{\timedtrace}{\tau}

\newcommand{\PO}{P}

\newcommand{\po}{PO\xspace}

\newcommand\reals{\mathbb{R}}

\DeclareMathOperator*{\argmin}{\arg \min}

\title{\LARGE \bf
Optimal  Planning for Timed Partial Order Specifications %\\
}

\author{Kandai Watanabe$^{1}$,
        Georgios Fainekos$^{2}$,
        Bardh Hoxha$^{2}$,
        Morteza Lahijanian$^{1}$,\\
        Hideki Okamoto$^{2}$,
        and
        Sriram Sankaranarayanan$^{1}$%
\thanks{$^{1}$Authors are with the Departments of Computer Science and Aerospace Engineering Sciences at University of Colorado Boulder, Boulder, Colorado. {\tt\small first.lastname@colorado.edu }}%
\thanks{$^{2}$Authors are with Toyota Motor North America, Research and Development. {\tt\small first.lastname@toyota.com}}%
}

\begin{document}
\AddToShipoutPictureBG*{%
  \AtPageUpperLeft{%
    \hspace{16.5cm}%
    \raisebox{-1.5cm}{%
      \makebox[0pt][r]{To appear in the 2024 IEEE International Conference on Robotics and Automation (ICRA2024), May 2024}}}}

\maketitle
\thispagestyle{empty}
\pagestyle{empty}

%%%%%%%%%%%%%%%%%%%%%%%%%%%%%%%%%%%%%%%%%%%%%%%%%%%%%%%%%%%%%%%%%%%%%%%%%%%%%%%%
\begin{abstract}
%Efficiency optimization in manufacturing processes is essential for businesses, with substantial financial implications. 
This paper addresses the challenge of planning a sequence of tasks to be performed by multiple robots while minimizing the overall completion time subject to timing and precedence constraints. Our approach uses the Timed Partial Orders (TPO) model to specify these constraints. We translate this problem into a Traveling Salesman Problem (TSP) variant with timing and precedent constraints, and we solve it as a Mixed Integer Linear Programming (MILP) problem. Our contributions include a general planning framework for TPO specifications, a MILP formulation accommodating time windows and precedent constraints, its extension to multi-robot scenarios, and a method to quantify plan robustness.  
We demonstrate our framework on several case studies, including an aircraft turnaround task involving three Jackal robots, highlighting the approach's potential applicability to important real-world problems.
Our benchmark results show that our MILP method outperforms state-of-the-art open-source TSP solvers OR-Tools.

% which show that MILP can be solved faster up to 80 nodes.
% , exemplifying its utility 
% Empirical experiments demonstrate significant computational efficiency gains 
% \td{R2: Compare to what?} \kw{Remove the previous sentence and add the next line. }, 
% enabling scalability. 
% The page limit is 6 pages for the paper (text, figures, tables, acknowledgement, etc.) + any number of pages for the bibliography/references. Papers exceeding the (6+n) page limit at the time of submission will be returned without review.
% \begin{itemize}
%     \item Paper Submission Deadline: 9/15
%     \item Video Submission Deadline: 9/17-21
%     \item Video Maximum duration: 180 sec.
% \end{itemize}
\end{abstract}

%%%%%%%%%%%%%%%%%%%%%%%%%%%%%%%%%%%%%%%%%%%%%%%%%%%%%%%%%%%%%%%%%%%%%%%%%%%%%%%%
\section{INTRODUCTION}\label{sec:introduction}

% In manufacturing factories/business processes, increasing efficiency is critical to their business outcome. 
% The change in efficiency could affect millions of dollars in the revenue of the company. 
% Thus, it is critical for companies to minimize the total time (a.k.a. makespan) of a workflow. 
% In a workflow, there are often times task ordering or timing constraints that must be satisfied to complete the task. 
% For example, water must be boiled before cooking pasta, or pasta must be cooked for 9-10 minutes. 
% These constraints are often partially pre-defined or can be obtained through mining from data \cite{watanabe2023timed, senderovich2019learning}.
% In this work, we are specifically interested in Timed Partial Order specifications (TPO) as they can succinctly capture a complex set of timing constraints, are extremely easy to specify, and can be learned easily from data \cite{watanabe2023timed}. 
% Additionally, we have spatial constraints such as traveling distances between locations in a floor map. 
% Some tasks cannot be executed in time, depending on the order. Thus, it is critical to schedule the tasks in the right order to be able to complete the task.
% In this work, we tackle the problem of synthesizing a plan that minimizes the total processing time while adhering to the given constraints. 

% Workflow analysis and optimization is of crucial importance to increase efficiency from manufacturing to administrative processes.   
% Workflows are governed by tasks that must be completed in a specific ordering and timing constraints. 

Workflow analysis and optimization techniques are of crucial importance in increasing efficiency across many domains, from manufacturing to administrative processes. These workflows are conventionally structured around tasks that have to be completed subject to precedence and timing constraints.   
% From the optimization is concise requirement representation and planning.  
% There exists a large body of work on representation and scheduling for such tasks.  
% For example, water must be boiled before cooking pasta, or pasta must be cooked for 9-10 minutes. 
Such constraints are often defined by the user or inferred from demonstrations~\cite{watanabe2023timed, senderovich2019learning}.
A recently-proposed model, \emph{Timed Partial Order} (TPO)~\cite{watanabe2023timed}, provides a succinct, understandable, and easily analyzable representation of these timing constraints. 
It allows for precedence constraints using partial orders, and the timing constraints are specified using clocks that can be reset when events in the workflow happen. % as various events in the workflow happen.  
However, the problem of planning task sequences given environmental constraints and TPO specifications has not been solved.  The main challenge lies in the combination of environmental constraints that specify how the events in the workflow can be achieved, timing costs for various events, and the  TPO. All of these are to be taken into account while planning for such agents.  
In this work, we develop a planning framework with TPO specifications against environments modeled as deterministic transition systems. 

\begin{figure}[t]
    \centering
    \begin{minipage}[t]{0.65\linewidth}
        \centering
        \includegraphics[width=\linewidth]{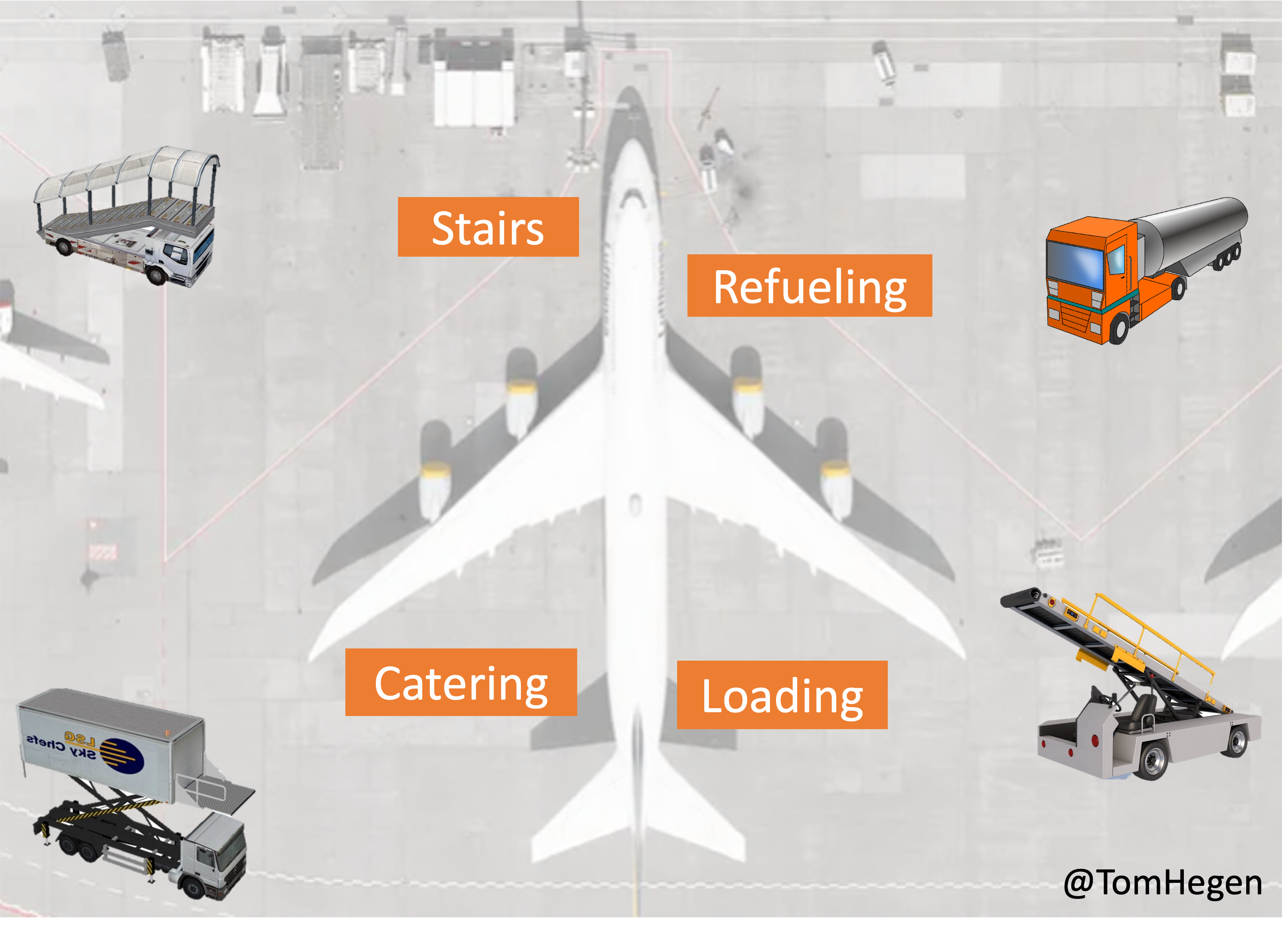}
        \subcaption{Top-down view}
        \label{fig:AircraftTurnaroundDepiction}
    \end{minipage}
    \begin{minipage}[t]{0.28\linewidth}
        \centering
        \includegraphics[width=\linewidth]{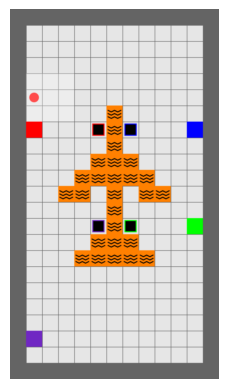}
        \subcaption{Gridworld}
        \label{fig:AircraftTurnaroundMinigrid}
    \end{minipage}
    \caption{Aircraft Turnaround Example}
    \label{fig:RunningExample}
    \vspace{-2mm}
\end{figure}

Consider the agent operating in an aircraft turnaround task in \Cref{fig:RunningExample}. Many tasks such as bulk loading/unloading and refueling can be completed in parallel. 
Additionally, there are precedence and timing constraints. 
For example, deplaning can only be done after the stair truck is placed, and deplaning takes at least $15$ minutes. 
Catering can only begin after all passengers have deplaned. 
To this end, we need a plan synthesis algorithm that can find the minimum-makespan plan under the timing and partial-order constraints.

% called "Mutli-Agent TIme-Constrained Planning Synthesis algorithm" ((auto)MATIC-PlanS) 
In this paper, we propose a new framework to solve the time-constrained plan synthesis problem for a single as well as multiple robots.  We employ TPOs as task specifications and use Deterministic Transition Systems (DTS) as abstraction models of the robots in their environments. We show that the planning problem on DTS with TPO specifications reduces to a type of Traveling Salesman Problem (TSP), which asks, given a map of cities, to find the lowest cost path to visit every city once. 
% \kw{To generalize the problem to visit every \textit{subsets} of the cities, we consider Generalized TSP.} 
TSP is well-studied and known to be NP-hard~\cite{Garey+Johnson/1979/Computers}, but there exist tools that use highly-effective heuristics, allowing fast computations~\cite{Cook+Others/2014/In}.
We formulate our DTS with TPO planning problem as an instance of TSP with the addition of timing and precedent constraints~\cite{parragh2007survey}. 
These constraints introduce the difficulty of applying off-the-shelf heuristics to our problem. Instead, we solve the problem as a Mixed Integer Linear Program (MILP), which finds an optimal solution rapidly for some of our benchmarks. We also show that for the multi-robot setting, a slight modification of the MILP can be employed. Furthermore, we provide an efficient approach to robustness analysis of the synthesized plans.
% The traveling times between cities (edge costs) can be computed by introducing the time measure between the events, e.g., the shortest distance between events. 
%%The benefits of our TSP formulation are 
%(1) simplicity (many planning problems can be transformed into a TSP), 
%(2) generality (heuristic approaches for solving TSP can be adapted to find approximate solutions), and 
%(3) extensibility (can be extended to different varieties of problems such as multi-agent TSP, different objective functions, pick-and-delivery, etc.).

The contributions of this work are fourfold: we 
(i) introduce a general planning framework for TPO specifications that can be applied to one or more robots, 
(ii) formulate a MILP with time windows (global time with respect to the start event) and precedence constraints (local time between sub-tasks), and extend it to multiple robots, 
(iii) propose a method based on LP to quantify the robustness of the synthesized plans to capture the lower and upper bounds on the delays that the plan can tolerate w.r.t. the given TPO, and 
(iv) provide a set of illustrative case studies and benchmarks that empirically show that TPO constraints actually narrow down the search space, speed up the computation time, and enable scaling up the algorithm to 160 nodes and 40 robots. We perform a physical experiment for an aircraft turnaround task with three Jackal robots that demonstrates the ability of our approach to plan for practically relevant problems.

% We evaluate our approach on simulations to show that the algorithm can find a time-efficient order that satisfies time and precedent constraints.
% We also show that our approach can scale up to more than 80 nodes (depending on the time windows), handle multiple duplicate nodes, and be extended to multi-agent systems without increasing the complexity of the problem. 

% In practice, users iteratively adjust their plans by adding constraints. We show that linear inequality timing constraints are easy to add, and our algorithm can easily handle such changes, which could be challenging for other decision techniques such as reinforcement learning. 

\section{RELATED WORK}\label{sec:related-work} 

Simple Temporal Networks (STNs) \cite{dechter1991temporal} and Timed Automata (TA) \cite{alur1994theory} allow us to specify complicated timing specifications. A comparison of TPOs with related formalisms is provided by Watanabe et al~\cite{watanabe2023timed}.

{\it Model Checking Problem \cite{alur1994theory}:} The problem focuses solely on properties such as consistency of the specification model, neglecting the operating environment. %It checks if the model is feasible, i.e., no contradictions exist. 
The focus of this paper involves plan synthesis which combines the timing specification and the model of the operating environment. 
% The model-checking problem is polynomial for STNs and it is PSPACE-complete for TAs.

{\it Plan Synthesis Problem:} The problem is to find a plan in the operating environment that satisfies the specification. This is a common problem in 
robotics, and many studies have been conducted for TAs \cite{asarin1998controller} and Signal Temporal Logic (STL) \cite{Maler:STL:2004} which can be translated into TAs \cite{MalerNP06}. 
In fact, the synthesis algorithm is known to be exponential in the number of clocks (Lemma 4.5 and Theorem 7.8 of \cite{alur1994theory}) and blows up very quickly in the number of tasks and environmental states.
This makes the algorithm difficult to apply to larger instances and multi-agent systems \cite{sun2022multi}. 
In this paper, we focus on TPOs that can be turned into linear inequality constraints, which reduce the complexity of the problem.

{\it Task \& Motion Planning Problem:} The problem is to assign (heterogeneous) tasks to robots, involving the problem of coalition formation \cite{guo2023recent}.
The work in \cite{suslova2020multi} tackles the task-allocation with timing and precedent constraints by forming coalitions of robots to accomplish tasks more efficiently. Similarly, \cite{gosrich2023multi} tackles the coalition formation problem to assign tasks to robots via a min-cost network flow approach. Our focus is not on allocating (heterogeneous) tasks to (heterogeneous) robots via forming coalitions, but rather it is on scheduling tasks that satisfy given formal specifications (e.g., precedence order, timing constraints, etc.). 
Others have also formulated the task allocation problem as a Traveling Salesman Problem (see  \cite{chakraa2023optimization} for a short survey), but without precedence and timing constraints.
% On the other hand, some literature considers precedence or timing constraints (see survey \cite{nunes2017taxonomy}) but in the context of heterogeneous robots and tasks which leads to algorithms and solutions which are more complicated than needed for the homogeneous robots problem. 
% , and taxonomy was of the similar problems were introduced in \cite{nunes2017taxonomy}. 
As highlighted in survey \cite{nunes2017taxonomy}, some works have considered timing and precedent constraints along with others (multi-agent, hard/soft constraints, deterministic/stochastic, etc.). 
However, no work has looked into local timing dependencies, and more importantly, into tasks that can be performed at multiple different locations.
The latter creates a choice of not only which robot should perform the task, but also in which location (for which we need to formulate a Generalized TSP).

% \cite{choudhury2022dynamic} solves online task allocation problem under uncertainty which is out of our scope.
% Surveys on GTSP \cite{pop2023comprehensive} and its one of the heuristics \cite{karapetyan2011lin}. 
% We picked a few related works and we classified them into \Cref{table:survey}.

% \begin{table}[t]
% \caption{Classification of the related work (Single-Agent)}
% \label{table:survey}
% \begin{tabular}{|l|llllll|}
% \hline
%           & \multicolumn{6}{l|}{Constraints}                                                                                   \\ \hline
%           & \multicolumn{2}{l|}{Time Window (TW)} & \multicolumn{2}{l|}{Precedence Relation (PR)} & \multicolumn{2}{l|}{TW-PC} \\ \hline
% Exact     & \multicolumn{2}{l|}{\cite{baker1983exact}}                 & \multicolumn{2}{l|}{\cite{ascheuer2000branch}}                         & \multicolumn{2}{l|}{\cite{mingozzi1997dynamic}}      \\ \hline
% Heuristic & \multicolumn{2}{l|}{\cite{bansal2004approximation, gendreau1998generalized, savelsbergh1985local, da2010general}}                 & \multicolumn{2}{l|}{\cite{gambardella1997has, escudero1988inexact}}                         & \multicolumn{2}{l|}{$\times$}      \\ \hline
% \end{tabular}
% \end{table}

\section{PROBLEM FORMULATION}\label{sec:problem-formulation}

% \begin{figure}[t]
% \begin{center}
% \begin{tikzpicture}
%     \node(n1) at (-2,0)[rectangle,draw=black, thin]{\footnotesize \begin{tabular}{c}
%     \textsc{Timed Partial Order}\\ 
%     (TPO)\\ 
%     \end{tabular}};
%      \node(n2) at (2,0)[rectangle, draw=black, thin]{\footnotesize \begin{tabular}{c}
%     \textsc{Det. Trs. System}\\ 
%     (DTS)\\
%     \end{tabular}};
% \node (n3) at (0,-1.5)[rectangle, draw=black, thin]{\footnotesize \begin{tabular}{c}
%    \textsc{Generalized TSP} + \\ 
%    \textsc{Time Windows} + \\ 
%    \textsc{Precedence}
%     \end{tabular}};
% \node (n4) at (0, -3)[rectangle, draw=black, thin]{\footnotesize \begin{tabular}{c}
% Plan Analysis\\ 
% \end{tabular}};
% \draw[line width=1.5pt, ->] (n1) edge (n3) 
% (n2) edge (n3)
% (n3) edge node[right]{\scriptsize Plan} (n4);
% \end{tikzpicture}
% \end{center}
% \caption{Overview of the approach in this paper: we combine a specification of timing constraints called timed partial orders with a deterministic transition system (DTS) that models various robot moves and their costs. These are integrated into a combinatorial optimization problem that yields plans which are them analyzed for various ``soft'' constraints such as robustness to timing jitters. }\label{fig:overview}
% \end{figure}

In this section, we first introduce TPOs and present the single/multi-robot problem formulations.
% Figure~\ref{fig:overview} shows an overview of the overall approach.

\subsection{Timed Partial Order (TPO) Specification}

TPOs provide a simple yet useful model for specifying timing constraints. It is closely related to   timed automata~\cite{Alur:IEEE:2000} and simple temporal networks (STNs)~\cite{dechter1991temporal}. We provide a brief summary of TPOs, referring  the reader to the paper by Watanabe et al for further details~\cite{watanabe2023timed}.

TPOs specify a timed sequence of \emph{events}. 
These events may include the start and finish of a given task, or an environmental event (e.g., the temperature of the water has exceeded 100$^\circ$C).  
% \td{Furthermore, we assume that repetitions of events are \emph{disambiguated} by giving them unique labels.} \kw{Repetitions induce a self-loop invalidating the important properties of the TPOs.}
The syntax of TPOs as defined in \cite{watanabe2023timed} does not allow an event to be repeated. Hence, we assume that repetitions of events are \emph{disambiguated} by giving them unique labels.
%Timing constraints
%specify the possible times at which these events may happen in a given run of a workflow.

Let $\events = \{\event_1, \ldots, \event_n \}$ be the set of events. A \emph{timed trace} $\timedtrace: \events \rightarrow \reals_{\geq 0}$  is a mapping 
$ \timedtrace= \{ \event_1 \mapsto t_1, \event_2 \mapsto t_2, \ldots, \event_n \mapsto  t_n \} $, 
wherein  $t_i \geq 0$ denotes the timestamp for event $\event_i \in \events$. %With an abuse of notation, we also use $\tau$ as a set.
Timed partial orders specify which timed traces are feasible (allowed) and which ones are not. 

%A timed-trace $\timedtrace$ induces an ordering over the events according to increasing time stamps, and thus may also be viewed as a sequence:
%$(\sigma_1, t^{(1)}) \ldots (\sigma_n, t^{(n)}),$ wherein each $\sigma_i \in \events$ denotes a unique $i^{th}$ event in the trace with corresponding time %stamp $t^{(i)}$ and furthermore, $t^{(1)} < t^{(2)} < \cdots < t^{(n)}$. %%
Recall that a \textit{(strict) Partial Order} (\po) $\PO$ is a relation $\prec$ on a set $\events$ that is irreflexive, asymmetric, and transitive. We write $e_i \preceq e_j$ if $e_i \prec e_j$ or $i = j$. If $\event_i \preceq \event_j$ holds, then for any timed trace $\timedtrace$ we require that $\tau(\event_i) \leq \tau(\event_j)$.

\begin{definition}[TPO]
A \emph{timed partial order} (TPO) is specified by a directed-acyclic graph (DAG) $\varphi: (\events, \prec)$ describing a strict partial order over $\events$ augmented with the following: 
\begin{compactenum}
    \item A finite set of \emph{clocks} $C = \{ c_1, \ldots, c_m \}$,
    \item A \emph{guard map} $g$
    that maps each event $\event_i$ to a guard condition, which is a conjunction of the form
$g(\event_i): \bigwedge_{j=1}^{n_i} c_j \bowtie a_j$, wherein $c_j \in C$ denotes a clock, $\bowtie \in \{ \leq, \geq \}$, and $a_j \in \reals_{\geq 0}$ is a non-negative constant, and
\item A \emph{reset map} $R: \events \to 2^{C}$ that associates each event $\event_i$ with a subset of clocks $R(\event_i) \subseteq C$ that are to be reset to $0$ whenever event $\event_i$ is encountered.
\end{compactenum}
\end{definition}

 We initialize all the clocks to $0$ and the clocks simply run to measure time. The edges in the TPO specify a precedence ordering between events. Thus, if the TPO has an edge $e_i \rightarrow e_j$, we require for any timed trace $\tau$ satisfying the TPO that $\tau(e_i) \leq \tau(e_j)$. Similarly, TPOs associate constraints over the clocks on each node. These constraints must hold whenever an event $e_i$ occurs and the reset actions corresponding to the events are executed to update the clock values.

%A valuation $\nu: C \rightarrow \reals_{\geq 0}$ assigns each clock $c_i \in C$ to a non-negative number $\nu(c_i)$. A given valuation $\nu$ can be advanced in time by a fixed  $\delta \geq 0$ to yield a new valuation  $\nu'\ :=\ \nu \oplus \delta$ such that $\nu'(c_j) = \nu(c_j) + \delta$ for all $c_j \in C$. Likewise, given a $\nu$ and a subset of clocks $\hat{C} \subseteq C$, we denote the valuation $\nu' := \textsf{reset}(\nu, \hat{C})$ as that obtained by setting each clock $c \in \hat{C} $ to be $0$:
%\begin{align*}
%\nu'(c) = \begin{cases}
%0 & c \in \hat{C} \\
%\nu(c) & c \not \in \hat{C} \end{cases}.
%\end{align*}

%Let $\nu_0$ represent a fixed special initial valuation wherein $\nu_0(c_j) = 0$ for all clocks and let $t^{(0)} = 0$. Timestamps represent global time since the inception of the process whereas clocks measure the time elapsed since their last reset.

\begin{figure}
\begin{center}

\begin{tikzpicture}
\begin{scope}
\matrix[every node/.style={draw=black, rectangle, rounded corners}, row sep=5pt, column sep=6pt]{
%           PlaceStairs -> Deboard -> Catering
% Arrival ->          Unload/Loading            -> MoveoutStairs
                & \node(n2){$e_2$}; & \node(n3) {$e_3$}; & \node(n4) {$e_4$}; \\
 \node(n1){$e_1$}; &  & & & \node(n6){$e_6$};\\
 & & \node(n5){$e_5$}; & \\
};
\path[->, line width=2pt] (n1) edge (n2)
(n2) edge (n3)
(n3) edge (n4)
(n4) edge (n6)
(n1) edge (n5)
(n5) edge (n6);

\draw (n1.south)++(0,-0.3) node{ \footnotesize $c_1 := 0$};
\draw (n2.north)++(0,0.3) node{ \footnotesize $c_2 := 0$};
\draw (n4.north)++(0,0.3) node{ \footnotesize $c_2 \geq 30$};

\draw (n6.south)++(0,-0.3) node{ \footnotesize $c_1 \leq 60$};
\end{scope}

\begin{scope}[xshift=4.2cm]
\node at (0,0) {
\begin{tabular}{ll}
\hline
$e_1$ & Airplane Arrived \\
$e_2$ & Place Stairs \\
$e_3$ & Deboarding \\
$e_4$ & Catering \\
$e_5$ & Unloading / Loading \\
$e_6$ & Move out Stairs \\
\hline
\end{tabular}
};
\end{scope}
\end{tikzpicture}
\end{center}
\vspace{-4mm}
\caption{Partial Order for an aircraft turnaround task. Events $e_1, \ldots, e_5$ represent events such as ``airplane arrived'' ($e_1$) or the commencement of various tasks such as ``Catering'' ($e_4$).}
\label{fig:RunningExampleTPO}
\vspace{-4mm}
\end{figure}
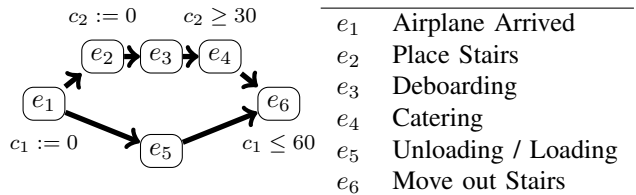

\begin{myexam}
    \Cref{fig:RunningExampleTPO} shows the TPO specification for the aircraft turnaround task depicted in \Cref{fig:RunningExample}. The TPO expresses many partial order constraints. 
    For example, stairs must be placed (event $e_2$) before passengers deboard event ($e_3$). After deboarding, the catering service can be started ($e_4$). 
    % Meanwhile, refueling can be started immediately after the vehicle is placed by the airplane, ends in $\leq 20$ time units, and the vehicles can be moved out. 
    Meanwhile, bulk unloading and loading ($e_5$) can be done in parallel.
    Only after everything is completed, the stairs can be removed ($e_6$). The clocks enforce additional timing constraints. For instance, clock $c_1$ is reset when we encounter event $e_1$ and event $e_6$ has the constraint $c_1 \leq 60$. This enforces the constraints that the stairs must be moved out within $60$ time units of the aircraft arriving. Likewise, another clock $c_2$ expresses the constraint that catering cannot be started until at least $30$ minutes after the plane arrives (to allow sufficiently many passengers time to de-board).
\end{myexam}

Watanabe et al \cite{watanabe2023timed} showed that TPOs $\varphi$ can be translated into a conjunction of the inequality forms, %With the abuse of the notation $\varphi$,
\begin{equation}
    \label{eq:TPO inquality}
    \varphi: \!\!\! \bigwedge_{i, j\ \mathsf{s.t.}\ e_i \prec e_j} \!\!\!\!\! (t_j - t_i) \in [\ell_{j,i}, u_{j,i} ] \ \land\ \bigwedge_{j=1}^n t_j \in [a_j, b_j ],
\end{equation}
wherein $\ell_{i,j} \geq 0, a_j \geq 0$ form lower bounds and $u_{j,i}, b_j \in \reals_{\geq 0} \cup \{ \infty \} $ are upper bounds that can be non-negative real numbers as well as $ +\infty$. The calculation of these bounds can be easily automated but is not explained further here. 
Specifying constraints as inequalities is a simple way to specify the relationship between events.

\subsection{Single Robot Setting}

% \begin{figure}[t]
%     \centering
%     \caption{DTS of the Figure xxx}
%     \label{fig:example-environment}
% \end{figure}

In this work, we consider both single and multi-robot cases.  For the single agent setting,
we assume the agent operates deterministically in an environment.
Abstractions can be made to represent it as a discrete Determinisitic Transition System (DTS), 
% Such an abstraction is commonly used in formal method approaches to robotics \cite{Lahijanian:AR-CRAS:2018,Hadas:ICRA:2007,Lahijanian:ICRA:2009}.
similar to \cite{Hadas:ICRA:2007,Lahijanian:AR-CRAS:2018,Lahijanian:ICRA:2009}.

\begin{definition}[DTS]
    \label{def:DTS}
    A single-robot \emph{deterministic transition system} (DTS) is a tuple $\mathcal{T} = (X, A, x_0, \delta_T, \Pi, L)$, where 
    \begin{compactitem}
        \item $X$ is a finite set of states, 
        \item $A$ is a finite set of controls or actions, 
        \item $x_0 \in X$ is the initial state,
        \item $\delta_T: X \times A \rightarrow X$ is the (partial) transition function, 
        \item $\Delta_T: X \times A \rightarrow \mathbb{R}_{\geq 0}$ is the transition duration function, 
        \item $L: X \rightarrow \Pi \cup \{\emptyset\}$ is a labeling function that maps each state to an event or an empty set. Without loss of generality, we assume $L(x_0) = \emptyset.$
    \end{compactitem}
\end{definition}

% For example, the agent (the chef) in \Cref{fig:example-environment} has to move to each location to collect the ingredients and cook at the designated locations, e.g., at the stove and cutting board. 

A \textit{plan} $\gamma = \gamma_0 \gamma_1 \ldots \gamma_{n-1}$ is a sequence of actions, where $\gamma_i \in A$ for all $0\leq i \leq n-1$.  
A \textit{valid plan} is plan $\gamma$ that respects the transition function $\delta_T$, i.e, $\delta_T(s_{i},\gamma_{i})$ exists for all $0\leq i\leq n-1$.  We denote the set of all valid plans by $\Gamma$.
By executing $\gamma \in \Gamma$, the robot generates a \textit{trajectory} $s^\gamma = s^\gamma_0 s^\gamma_1 \ldots s^\gamma_n$, where $s^\gamma_0 = x_0$ and $s^\gamma_{i+1} = \delta_T(s^\gamma_{i},\gamma_{i})$.
The observation trace of a trajectory is the sequence of observed labels, i.e., $\rho^\gamma = L(s^\gamma_0)L(s^\gamma_1)\ldots L(s^\gamma_n)$.
The duration of a trajectory is a sum of the transition durations $D(s^\gamma) = \sum_{i=0}^{n-1} \Delta_T(s_i^\gamma, \gamma_i)$. 
This induces a timed trace $\tau^\gamma = (L(s^\gamma_0), t_0) \ldots (L(s^\gamma_n), t_n)$ where $t_i = D(s^\gamma_0,\ldots,s^\gamma_i)$,
which is equivalent to the definition of the timed trace in TPO.
We say plan $\gamma \in \Gamma$ satisfies a TPO if its timed trace $\tau^\gamma$ satisfies the inequalities $\varphi$ in~\eqref{eq:TPO inquality}, denoted by $\tau^\gamma \models \varphi$. \

\begin{myexam}
    Again consider the aircraft turnout example in \Cref{fig:AircraftTurnaroundMinigrid} with the red agent in position coordinate $(1,17)$. It can be modeled as a transition system, where each cell of the grid is associated with a state $x \in X$. The agent can take actions $A = \{up, \, down, \, right, \, le\!ft, \}$. 
    % Starting from $x_0$ with coordinate position $(1, 17)$, 
    Then, plan  $\gamma = down \, down$ generates trajectory $s^\gamma = (1,17)(1,16)(1,15)$, which induces timed trace $\tau^\gamma = (\emptyset,0)(\emptyset,1)(e_\text{red\_floor},2)$. 
\end{myexam}

Note that DTS can model a variety of different systems including robotic manipulators~\cite{He:ICRA:2015,He:RAL:2019}. 
% Now, the problem is how can the agent optimally plan such that the total processing time (makespan) is minimized while respecting the TPO constraints
% \ml{minimizing processing time? Why not the cost?} \kw{We can include the cost, but we still need to associate DTS and TPOs with time in order to check if the timing constraints in the TPO were satisfied. My plan is to show how the makespan can be minimized at first and later I was planning to extend the "framework" to different objectives. Minimizing makespan is generally the most likely thing that people in operational research are interested in. For more detail, please read the book "Scheduling theory, algorithms" (the link is commented out).}
% https://doc.lagout.org/science/0_Computer%20Science/2_Algorithms/Scheduling_%20Theory%2C%20Algorithms%2C%20and%20Systems%20%285th%20ed.%29%20%5BPinedo%202016-02-11%5D.pdf
%The problem is to find a plan with a minimum time duration that satisfies TPO.

\begin{problem}(Single Robot Plan Synthesis)\label{problem:single-agent} 
    Given a robotic system as a DTS $\mathcal{T}$ with a specification as a TPO $\varphi$, synthesize a valid plan $\gamma^* \in \Gamma$ for the robot that satisfies the TPO in a minimum time duration, i.e., 
    \begin{align*}
        \gamma^* = \arg\min_{\gamma \in \Gamma} D(s^\gamma) \quad \text{s.t.} \quad \tau^\gamma \models \varphi
    \end{align*}
\end{problem}

We further extend the problem to a multi-agent system. 

\subsection{Multi-Robot Setting}
We consider the same environment as above but now with $p\in \mathbb{N}_{> 1}$ robots, each with its own initial state.  
We define a multi-robot DTS (mDTS) by extending $\mathcal{T}$ to have a set of initial states $X_I = \{x_0^1, x_0^2, \ldots, x_0^p\} \subseteq X$, i.e., $\mathcal{T}^M = (X, A, X_I, \delta_T, \Pi, L)$, where $X,A,\delta_T, \Pi$, and $L$ are as in Def.~\ref{def:DTS}. 
% Note that transition function $\delta_T$ can also be viewed as transition relation $\delta_T \subseteq X \times A \times X$, where $\delta_T($
The notion of valid plan $\gamma^i \in \Gamma^i$ for robot $i$ is adopted from the single case, and the set of all valid plans for all robots is denoted by $\Gamma = \bigcup_{i=1}^p \Gamma^i$.

Similar to the single robot case, from initial state $x_0^i \in X_I$, plan $\gamma^i$ induces a trajectory $s^{\gamma^i}\!$, and a timed trace $\tau^{\gamma^i}$, and timetamps $t^{\gamma^i}$\!\!.
We assume that when a robot executes action $a \in A$ at state $x \in X$, it remains in $x$ for the entire duration $\Delta_T(x,a)$ before transitioning to $x' = \delta_T(x,a)$.  Then, the induced trajectory can be viewed as a piecewise function of time.  With an abuse of notation, we use $s^{\gamma^i}: \mathbb{R}_{\geq 0} \to X$ to denote this function, where $s^{\gamma^i}\!(t)$ is the state visited by trajectory $s^{\gamma^i}$ at time $t$.
We say two trajectories $s^{\gamma^1}$ and $s^{\gamma^2}$ are \textit{non-colliding} if, for all $t \leq \max\{D(s^{\gamma^1}),D(s^{\gamma^2}) \}$, $s^{\gamma^1}\!(t) \neq s^{\gamma^2}\!(t)$.
% if the following conditions satisfy for all steps $1 \leq k < m$. 
% \begin{enumerate}
%     \item $u_k \neq v_k$ (i.e., no location collisions),
%     \item $(u_k, u_{k+1}) \neq (v_k, v_{k+1})$ (i.e., no edge collisions),
% \end{enumerate}
% if $u_k \neq v_k$ for all steps $1 \leq k < m$, i.e., no location collisions.
% if $u_k \neq v_k$ for all steps $1 \leq k < m$, i.e., no location collisions.
We define a timed trace $\tau^{\gamma^{1}\ldots \gamma^{p}}$ of the multi-robot system to be the  union of the individual robot's timed traces. A timed trace is also viewed as a set with the order relation induced by the timestamps.

Then, the multi-robot problem is to find plans for the $p$ robots that generate non-colliding trajectories with the timed trace $\tau^{\gamma^{1},\ldots, \gamma^{p}}$ that satisfies the TPO specification with a minimum time duration.

\begin{problem}[Multi-Robot Plan Synthesis]
    \label{problem:multi-agent}
    Given a system of $p$ robots as a mDTS $\mathcal{T}^M$ and a TPO specification $\varphi$, synthesize plans $\gamma^{1*},\ldots,\gamma^{p*} \in \Gamma$ under which the multi-robot system satisfies the TPO in a minimum time duration:
    \begin{align*}
        &\gamma^{1*},\ldots, \gamma^{p*} = \argmin_{\gamma^1...\gamma^p \in \Gamma} \max \{D(s^{\gamma^1}), \ldots, D(s^{\gamma^p})\} \\
        &\text{subject to}\\ 
        &\qquad\quad \tau^{\gamma^{1} \ldots \gamma^{p}} \models \varphi \\
        &\qquad\quad s^{\gamma^j} \ \text{and}\ s^{\gamma^j} \text{are non-colliding} \quad \forall \gamma^i, \gamma^j \in \Gamma.
    \end{align*}
\end{problem}

Note that Problems~\ref{problem:single-agent} and ~\ref{problem:multi-agent} ask for a satisfying plan that minimizes the maximum duration of the induced trajectories, which is also known as the \textit{makespan} of the plan.

% A naive approach to this problem that can be applied out-of-the-box which we show in the experiment section is the Product Graph-based shortest path planning approach commonly used in the formal method community, i.e., take the composition of the specification graph and an environment graph and finding the shortest path on the graph (Refer to Hadas' and Morteza's Survey Paper?)
% However, the size of the graph could exponentially increase as the number of events and timing constraints increases because converting TPOs to Timed Automata could potentially be exponential to the number of events. Model-checking of Timed Automata requires them to be converted to a model called Zone-Automata which is exponential to the number of clocks.
% Therefore, we must seek an alternative approach to quickly find a solution to this problem. 

\section{Approach}\label{sec:approach}

In this section, we explain how a \Cref{problem:single-agent} can be formulated as an instance of the Generalized Traveling Salesman Problem (GTSP), but with timing and precedent constraints. These additional constraints introduce difficulty in applying existing heuristics to our problem out of the box. In this paper, we first focus on the Mixed Integer Linear Program (MILP) formulation.
We first introduce the problem of GTSP and later discuss how the problem can be translated into the graph representation of GTSP.  

\subsection{\Cref{problem:single-agent} as Generalized Traveling Salesman Problem}
The Generalized Traveling Salesman Problem \cite{noon1988generalized} is the problem of finding the minimum cost path that visits exactly one city from given subsets of cities.
GTSP is known to be an NP-hard problem, and it is a well-studied problem in combinatorial optimization research.
There are many existing methods to obtain either exact or approximate solutions to this problem \cite{pop2023comprehensive}. 
% The exact solution approach is suited for small instances of the problem, and the approximate solution approach is suited for larger instances of the problem with over XXX clusters. 
We want to translate \Cref{problem:single-agent} into a GTSP, more specifically, GTSP with Time-Windows and Precedence Relations (GTSP-TWPR) \cite{mingozzi1997dynamic} so that we can utilize the existing approaches and extend the problem formulation.
Formally, GTSP-TWPR is defined as follows.

\begin{definition}[GTSP-TWPR]
    Let $G=(V, E)$ be a weighted directed graph with vertices $V$ and edges $E = V\times V$.
    Node $v_i \in V$ is associated with a vertex cost $d_i$, which represents the time delay at that vertex.
    Edge $(v_i,v_j) \in E$ is assigned a time cost $d_{ij}$, which is the time required to move from $v_i$ to $v_j$. 
    % Additionally, 
    % and the edge cost $d_{ij}$ as the time required to move from node $v_i$ to $v_j$.  
    Node $v_0 \in V$ is designated as the depot with $d_0 = 0$.
    
    The problem seeks a tour that visits some of the vertices $v_{i_0}, v_{i_1}, \ldots, v_{i_m}$ with starting and ending at the depot, $v_{i_0} = v_{i_m} = v_0$, while minimizing the total time of the tour: $\sum_{j=0}^m d_{i_j} + d_{i_j, i_{j+1}}$. Note that we can set $d_{i,0} = 0$ for all $i$ if return to the depot is not required for the problem. We refer to this total time as the \emph{makespan} of the tour. The tour is subject to the following additional constraints: 
    (a) We partition the set $V \cup \{v_0\}$ 
    into disjoint subsets $V_1, V_2, \ldots, V_k$
    . The TSP tour is required to visit exactly one node from each subset $V_i$.  Let $t_i$ be the time at which the node in the set $V_i$ is visited by our tour. 
    (b) We require that $l_i \leq t_i \leq u_i$ for a time interval $[l_i, u_i]$ provided as input. (c)  We specify qualitative precedence constraints of the form $V_i \prec V_j$ that specifies that the tour must visit a node in $V_i$ before it visits some node in $V_j$. (d) Whenever we have $V_i \prec V_j$, we require $t_i \leq t_j$. Additionally, we may also  specify quantitative relative time window $[l_{ij}, u_{ij}]$ requiring that $l_{ij} \leq t_j - t_i \leq u_{ij}$.
    
\end{definition}

We show how \Cref{problem:single-agent} is  mapped to a GTSP-TWPR. %Our core idea is to treat TPO events as subsets of nodes. Each event must occur exactly once, which is equivalent to visiting each subset exactly once. 
% Nodes represent the states in the DTS whose labels are associated with the TPO events and, edge cost can be replaced with the cost between these states. 

\begin{definition}[Translation of \Cref{problem:single-agent} to GTSP-TWPC]
We abstract DTS $\mathcal{T}$ and the TPO $\varphi$ to a GTSP-TWPR graph, by defining the set of node $V = \{x_0\} \cup \{x \in X \mid L(x) \neq \emptyset \}$ to be the set of states with non-empty labels as well as $x_0$.  Then, the depot node $v_0 = x_0$, and
subset $V_i$ is a set of states whose label is event $e_i$, i.e., $V_i = \{x \in X | L(x) = e_i \}$.
% \ml{delete from here}
% Nodes $V \subseteq X$ are the union of all the sets $V = \{v_j \in V_i | V_i \in \vec{V} \} \cup \{v_0\}$, wherein $v_0 = x_0$ is the depot.
% \ml{to here}
A directed edge $(x_i, x_j)$ is added whenever $L(x_i) \preceq L(x_j)$ or the events $L(x_i), L(x_j)$ can happen in parallel.
The edge cost $d_{ij}$ between two states $x_i$ and $x_j$ is defined by the trajectory duration $D(s, \gamma)$ where $s$ is the shortest path between $x_i$ and $x_j$ on $\mathcal{T}$, i.e., $s = s_0 s_1 \ldots s_n$ that induces a trace $L(x_i)\emptyset... \emptyset L(x_j)$. Likewise, if state $x_i$  in $T$ has a self-transition under action $a$ with time cost $\Delta_T(x_i,a)$, then we set $d_i = \Delta_T(x_i,a)$ as the vertex cost for node $x_i$; otherwise, we set $d_i = 0$.
% TPO constraints are directly translated into time window and precedent constraints, i.e., $\bigwedge_{e_i \prec e_j} t_j - t_i \geq a_{ij}$ and $t_i \prec t_j$. 
\end{definition}

Edge costs are calculated by running an all-shortest path algorithms such as Floyd-Warshall algorithm \cite{floyd1962algorithm}. 
We note that our method can be extend to a continuous space kinodynamical robots by running Stable Sparse RRT (SST) \cite{li2016asymptotically} to obtain the shortest paths between states. 
Generally, this is run once for environments where locations are fixed, e.g., manufacturing factories, hospitals, etc.
\newcommand\comment[1]{}

\comment{
\begin{figure}[t]
    \centering
    \begin{subfigure}[b]{0.49\linewidth}
        \scalebox{0.85}{\begin{tikzpicture}[auto, node distance=1.3cm, every node/.style={rectangle, rounded corners}]

  % \node[label={[yshift=3mm]left:$\tau_0$}, 
        % label={[xshift=3mm]below:$\tau_{N+1}$}] (node0) {$v_0$};
  \node[draw] (node0) {$v_0$};
  % Cluster 1
  \node[draw, fill=blue!20, above of=node0] (node1) {$v_1$};
  \node[draw, fill=blue!20, right of=node1] (node2) {$v_2$};
  % Cluster 2
  \node[draw, fill=red!20, right of=node0] (node3) {$v_3$};
  % Cluster 3
  \node[draw, fill=green!20, above right of=node3] (node4) {$v_4$};
  \node[draw, fill=green!20, below right of=node4] (node5) {$v_5$};

  % Oval shapes to represent clusters
  \node[draw, thick, fit=(node0), label={[xshift=-0.5cm, yshift=-1.3cm]$V_0$}] {};
  \node[draw, thick, fit=(node1)(node2), label=$V_1$] {};
  \node[draw, thick, fit=(node3), label={[yshift=-1.3cm]$V_2$}] {};
  \node[draw, thick, fit=(node4)(node5), label=$V_3$] {};

  % \path[->, line width=1pt] (node0) edge node[left]{$y_{0,1}$} (node1)
  %                           (node1) edge node[left]{$y_{1,5}$} (node5)
  %                           (node5) edge node[above, yshift=-1mm]{$y_{5,3}$} (node3)
  %                           (node3) edge node[above, yshift=-1mm]{$y_{3,0}$} (node0);
  \path[->, line width=1pt] (node0) edge (node1)
                            (node1) edge (node5)
                            (node5) edge (node3)
                            (node3) edge (node0);

\end{tikzpicture}}
        \caption{Single Agent}
        \label{fig:GTSPSingle}
    \end{subfigure}
    \begin{subfigure}[b]{0.49\linewidth}
        \centering
        \scalebox{0.85}{\begin{tikzpicture}[auto, node distance=1.3cm, every node/.style={draw=black, rectangle, rounded corners}]
  % Initial State Cluster
  \node[] (node11) {$v_0^1$};
  \node[above of=node11] (node12) {$v_0^2$};
  \node[right of=node11] (node13) {$v_0^3$};
  % Cluster 1
  \node[fill=blue!20, above of=node12] (node1) {$v_1$};
  \node[fill=blue!20, right of=node1] (node2) {$v_2$};
  % Cluster 2
  \node[fill=red!20, below right of=node2] (node3) {$v_3$};
  % Cluster 3
  \node[fill=green!20, above right of=node3] (node4) {$v_4$};
  \node[fill=green!20, below right of=node4] (node5) {$v_5$};
  % Cluster 4
  \node[fill=blue!20, right of=node13] (node6) {$v_6$};
  \node[fill=blue!20, right of=node6] (node7) {$v_7$};
  \node[fill=blue!20, above right of=node6] (node8) {$v_8$};
  % Cluster 5
  \node[fill=yellow!20, below of=node2] (node9) {$v_9$};
  % Oval shapes to represent clusters
  \node[draw, thick, fit=(node1)(node2), label=$V_1$] {};
  \node[draw, thick, fit=(node3), label=$V_2$] {};
  \node[draw, thick, fit=(node4)(node5), label=$V_3$] {};
  \node[draw, thick, fit=(node6)(node7)(node8), label={[yshift=-2.3cm]$V_4$}] {};
  \node[draw, thick, fit=(node9), label=$V_5$] {};
  \node[draw, thick, fit=(node11), label={[yshift=-1.5cm]$V_0^1$}] {};
  \node[draw, thick, fit=(node12), label={[yshift=-1.5cm]$V_0^2$}] {};
  \node[draw, thick, fit=(node13), label={[yshift=-1.5cm]$V_0^3$}] {};

  \path[->, line width=1pt] (node12) edge[bend left] (node2)
                            (node2) edge[bend left] (node12)
                            (node11) edge (node9)
                            (node9) edge (node3)
                            (node3) edge (node5)
                            (node5) edge (node11)
                            (node13) edge[bend left] (node6)
                            (node6) edge[bend left] (node13);
\end{tikzpicture}}
        \caption{Multi Agents}
        \label{fig:GTSPMulti}
    \end{subfigure}
\caption{Depiction of the GTSP. The nodes represent the TSP nodes and the clusters represent the set. A black solid line represents a tour (tours).}
\label{fig:GTSPDepiction}
\end{figure}
}

\subsection{MILP Formulation}
% , and further introduces the heuristic approach to speed up the search for an approximate solution such as Ant Colony Optimization \cite{gambardella1997has}. 

\subsubsection{Single Robot}
We solve Problem~\ref{problem:single-agent} on graph $G$ exactly using a Mixed Integer Linear Programming (MILP) formulation. The formulation is shown in \Cref{fig:milp-formulation}.
Recall that $n = |\Pi|$ is the number of events, and per our construction of $G$, it is also the number of the subsets $V_1, \ldots, V_n$. We define the square bracket $[k]$ to represent the set $\{1,\ldots, k\}$. Let $N$ be the number of nodes $N=|V|$, 
$y_{ij}$ be an integer variable indicating the active edge $i\rightarrow j$, and $\tau_i$ be the continuous time variable that represents the completion of the $i$th event, where $\tau_{N+1}$ represents the time coming back to the depot. 
We additionally introduce the set $V_0=\{v_0\}$ with the depot node and the indices $I=\{0\}$ to simplify the formulation.
% \begin{figure}[t]
% \begin{equation*}
% \begin{aligned}
%     \min \quad & \tau_{N+1}\\
%     \textrm{s.t.} \quad & \sum_i y_{i,j} - \sum_k y_{j,k} = 0, & j\in [0,N]\\
%                     & \sum_{v_j \in V_l} \sum_i y_{i,j} = 1, & l \in [0,n] \\
%                     & \sum_{v_j \in V_l} \sum_k y_{j,k} = 1, & l \in [0,n] \\
%                     & \tau_j - \tau_i \bowtie a_{i,j}, & t_j, t_i, a_{i,j} \in \varphi  \\
%                     & y_{i,j}=1 \implies \tau_j - \tau_i \geq d_{i,j} + d_j, & i\in [0,N], \\
%                     & & j \in [1,N] \\[3pt]
%                     & y_{i,0}=1 \implies \tau_{N+1} - \tau_i \geq d_{i,0}, & i\in [1, N] \\
%                     & y_{i,j} \in \{0, 1\}, & i\neq j \in [0,N] \\
%                     & \tau_i \geq 0, & i \in [0, N+1]
% \end{aligned}
% \end{equation*}
% \caption{MILP formulation of the GTSP-TWPR problem.}\label{fig:milp-formulation}
% \end{figure}

Note that the continuous variables include $N+1^\text{th}$ variable $\tau_{N+1}$ that represents the time at the end of the tour.
Constraint \eqref{eq:milp1} expresses that the number of incoming and outgoing edges must be equal at every node. 
Constraints \eqref{eq:milp2} and \eqref{eq:milp3} represent that there is only one incoming and one outgoing edge for each subset. Together with the first constraint, we ensure that the incoming/outgoing edge to a particular subset $V_i$ must involve the same node $v_i \in V_i$.
Constraint \eqref{eq:milp4} represents all the TPO constraints, and 
\eqref{eq:milp5} delays the $j^\mathrm{th}$ event by $d_{i,j} + d_j$ from $i^\mathrm{th}$ event only if the edge is activated ($y_{i,j}=1$). 
Constraint \eqref{eq:milp6} delays $N+1^\text{th}$ visit (makespan) by the edge cost between $[N]$ nodes back to the initial nodes $I$.
$\implies$ represents ``implies" and can be expressed by using the Big-M method \cite{williams2013model}. 

A tour can be obtained by following the enabled edges $y_{i,j}=1$ from the depot node.
%This solution is an optimal tour that satisfies the TPO specification in the minimum makespan.
%Once the tour is obtained, a plan can be computed by backtracking the shortest paths between states. 

\begin{figure}[t]
\begin{subequations}
\begin{align}
    &\min \ \tau_{N+1} \quad  \textrm{s.t.} & \nonumber \\ %\label{eq:milp0} \\
    &\ \sum_i y_{i,j} - \sum_k y_{j,k} = 0, & j\in I\cup[N] \label{eq:milp1}\\
    &\ \sum_{v_j \in V_l} \sum_i y_{i,j} = 1, & l \in I\cup[n] \label{eq:milp2}\\
    &\ \sum_{v_j \in V_l} \sum_k y_{j,k} = 1, & l \in I\cup[n] \label{eq:milp3}\\
    &\ \tau_j - \tau_i \bowtie a_{i,j}, & t_j, t_i, a_{i,j} \in \varphi \label{eq:milp4}\\
    &\ y_{i,j}=1 \implies \tau_j - \tau_i \geq d_{i,j} + d_j, & i\in I\cup[N], \nonumber \\
                    &\ & j \in [N] \label{eq:milp5}\\
    &\ y_{i,j}=1 \implies \tau_{N+1} - \tau_i \geq d_{i,j}, & i\in [N], j\in I\label{eq:milp6}\\
    &\ y_{i,j} \in \{0, 1\}, & i, j \in I\cup[N] \\
    &\ \tau_i \geq 0, & i \in I\cup[N+1]
\end{align}
\end{subequations}
\vspace{-5mm}
\caption{MILP formulation of the GTSP-TWPR problem. Notation $[N] = \{1, \ldots, N\}.$}\label{fig:milp-formulation}
\vspace{-3mm}
\end{figure}

% Since some constraints are known beforehand, we can further reduce the decision variables from the original MILP formulation. 
% \begin{enumerate}
%     \item Precedent Constraint $t_i \leq t_j$: The integer variable $x_{ji}$ can be set to $0$.
%     \item Timing Constraints $t_j - t_i \geq 0$: Similarly, $x_{ji}$ can be set to $0$ and $t_j - t_i \geq 0$ can be added to reduce the search space.
% \end{enumerate}

% \subsection{MILP Formulation for Multi-Robots}
\subsubsection{Multiple Robots}

For the multi-robot setting, we construct graph $G$ in a similar manner as the single-robot case, except that for each initial state $x^i_0 \in X_0$, we add a depot node $v^i_0$ and its associated subset $V^i_0$ to $G$.
% Then, to solve Problem~\ref{problem:multi-agent}, we can extend the above MILP to multi-agent formulation, as depicted in \Cref{fig:GTSPMulti}. 
For simplicity, we redefine the indices $I=\{0_1, \ldots, 0_p\}$ to denote the depot nodes and their subsets.  
The MILP formulation then becomes exactly the same as that of the single-agent case in \Cref{fig:milp-formulation}. 
In practice, we avoided introducing new depots, but instead, we changed the number of incoming and outgoing edges at the initial node $v_0$ to prevent the increase in the number of integer variables. 
The right-hand sides of \eqref{eq:milp2} and \eqref{eq:milp3} equate to the number of robots starting at $V_0$.
% This implies that the complexity does not increase as much from that of the single-agent formulation.  This makes the approach scalable with respect to the number of robots, unlike the traditional multi-agent planning frameworks, where the complexity increases exponentially with the number of agents.

% \begin{equation}
% \begin{aligned}
%     \min \quad & \tau_{n+1}\\
%     \textrm{s.t.} \quad & \sum_i y_{i,j} - \sum_k y_{j,k} = 0, & j=1, \ldots, n\\
%                     & \sum_{v_j \in V_l} \sum_i y_{1,j} = p, & l=2, \ldots, n \\
%                     & \sum_{v_j \in V_l} \sum_k y_{j,1} = p, & l=2, \ldots, n \\
%                     & \sum_{v_j \in V_l} \sum_i y_{i,j} = 1, & l=2, \ldots, n \\
%                     & \sum_{v_j \in V_l} \sum_k y_{j,k} = 1, & l=2, \ldots, n \\
%                     & \tau_j - \tau_i \bowtie a_{i,j}, & \forall i,j \in \prec \\
%                     & y_{i,j}==1 \implies \tau_j - \tau_i \geq d_{i,j} + d_j, & 1 \leq i\neq j \leq n\\
%                     & y_{i,1}==1 \implies \tau_{n+1} - \tau_i \geq d_{i,1}, & i=2, \ldots, n\\
%                     & y_{i,j} \in \{0, 1\}, & 1 \leq i\neq j \leq n \\
%                     & \tau_i \geq 0, & i=1, \ldots, n+1 
% \end{aligned}
% \end{equation}

%Similar to the single-agent case, we can obtain a tour for each robot by following the enabled edges $y_{i,j}=1$ starting from and ending at its depot node. 
The resulting tours minimize the makespan, and their induced timed trace satisfies the TPO specification. 
%%These tours can be mapped to plans on $\mathcal{T}^M$ by backtracking the shortest paths, similar to the single robot case. 
However, when the tours are mapped to plans on  $\mathcal{T}^M$, they are not guaranteed to produce non-colliding trajectories.  Hence, they need to be check for collisions as a post process.  If a collision is detected, then we repair the plans by introducing delays to the individual plans as in \cite{Kottinger:AAAI:2024}. Acceptable limits on these delays can be calculated using a robustness analysis. 
Another approach is to resolve conflicts via the Conflict-Based Search (CBS) method for multi-agent as in \cite{ren2022conflict}. 
%We are currently investigating approaches to eliminate collisions  incrementally by adding incrementally adding constraints that prevent a colliding path from the set of remaining solutions.

\subsection{Robustness Analysis}

Once a tour is returned by our MILP formalism, it is now possible to understand how robust the tour is to variations in the edge and node costs or in other words, variations in the delays associated with a given node or an edge between two locations. Such delays is common during plan execution. Consider a single agent tour with edge costs: 
$v_0 \xrightarrow{d_{01}} v_1 \xrightarrow{d_{12}} v_2 \cdots v_{m-1} \xrightarrow{d_{m-1, 0}} v_0 $.
Let $d_i$ be the node cost of $v_i$ with $d_0 = 0$. Our goal is to characterize all possible timing variations $\delta(d_{ij})$ to the edge costs and $\delta(d_j)$ to the node costs such that the TPO constraints will continue to hold. To this end, note that the nominal
visit time for node $v_i$ in the tour is given by 
$t_i = \sum_{j=0}^{i-1} \Bigl( d_j + d_{j,j+1} \Bigr)$ for $i \in [1, m-1]$ while the visit time taking into account the unknown variations in $d_j$, $d_{i,j}$ will be
$ t_i = \sum_{j=0}^{i-1} \Bigl( d_j + d_{j,j+1} + \delta(d_j)  +\delta(d_{j, j+1}) \Bigr)$. Robustness analysis seeks a uniform bound $\epsilon$ such that whenever each $|\delta(d_j)| \leq \epsilon$ and $|\delta(d_{i,j})| \leq \epsilon$, the TPO constraints are guaranteed to hold. To compute such a limit $\epsilon$, we formulate the LP: 
% $ t_i = \sum_{j=0}^{i-1} \Bigl( d_j + d_{j,j+1} + \delta(d_j)  +\delta(d_{j, j+1}) \Bigr)$. Robustness analysis seeks a uniform bound $\epsilon$ such that whenever each $\epsilon \leq \delta(d_j)$ and $\epsilon \leq \delta(d_{i,j})$, the TPO constraints are guaranteed to hold. To compute such a limit $\epsilon$, we formulate the LP: 
% \vspace{-1mm}
\[ \begin{array}{rll}
\max & \epsilon \\
\text{s.t.} & t_i = \sum_{j=0}^{i-1} \big( d_j + d_{j,j+1} + \delta(d_j)  +\delta(d_{j, j+1}) \big)  \\
& \hspace*{2.5cm}\ \text{for}\ i = 1, \ldots, m-1 \\[2pt]
& t_i  - t_j \bowtie a_{i,j}, \; \; \text{ TPO constraints}\ t_i, d_j, a_{i,j}\ \\ 
% & -\epsilon \ \leq\ \delta(d_j) \ \leq\ \epsilon \\ 
% & -\epsilon\ \leq\ \delta(d_{i,j})\ \leq\ \epsilon \\ 
% & \epsilon \leq t_j - t_i \\
& \epsilon \ \leq\ \delta(d_j)\\ 
& \epsilon\ \leq\ \delta(d_{i,j})\\ 
\end{array}\]

The LP above is always feasible and its optimal solution $\epsilon$ denotes a uniform bound on the timing variations $\delta(d_j), \delta(d_{i,j})$ such that as long as $|\delta(d_j)| \leq \epsilon$ and 
$|\delta(d_{i,j})| \leq \epsilon$ for each node and edge delay, the tour continues to be a feasible solution that satisfies the TPO constraints.  The formulation for a multi-robot tour is almost identical except that the times $t_i$ are computed differently for each robot. Unfortunately, trying to incorporate the robustness analysis as part of the solution to the TSP itself results in a  robust optimization problem which often requires more  expensive approaches to solve. Our future work will consider incrementally eliminating tours that fail a robustness criterion by adding a ``blocking'' constraint to force the MILP solver to return back with a different tour. 

% \subsection{\kw{Complexity of other approaches}}

% \subsection{\kw{Extensions to the current formulation}}
% Mention that this can be extended to other objectives commonly used in operational research. For example, makespan, sum of each task time(?), violations, number of violations, and etc.... Show that the algorithm can add other constraints. For example, the cost limitation ($\sum d_{ij}\cdot x_{ij} \leq D$) or resource constraints. 

% See \cite{parragh2007survey}

\section{Experiments}\label{sec:experiments}

\begin{figure}[b]
    \centering
    \begin{minipage}[t]{0.47\linewidth}
        \centering
        \includegraphics[width=\linewidth]{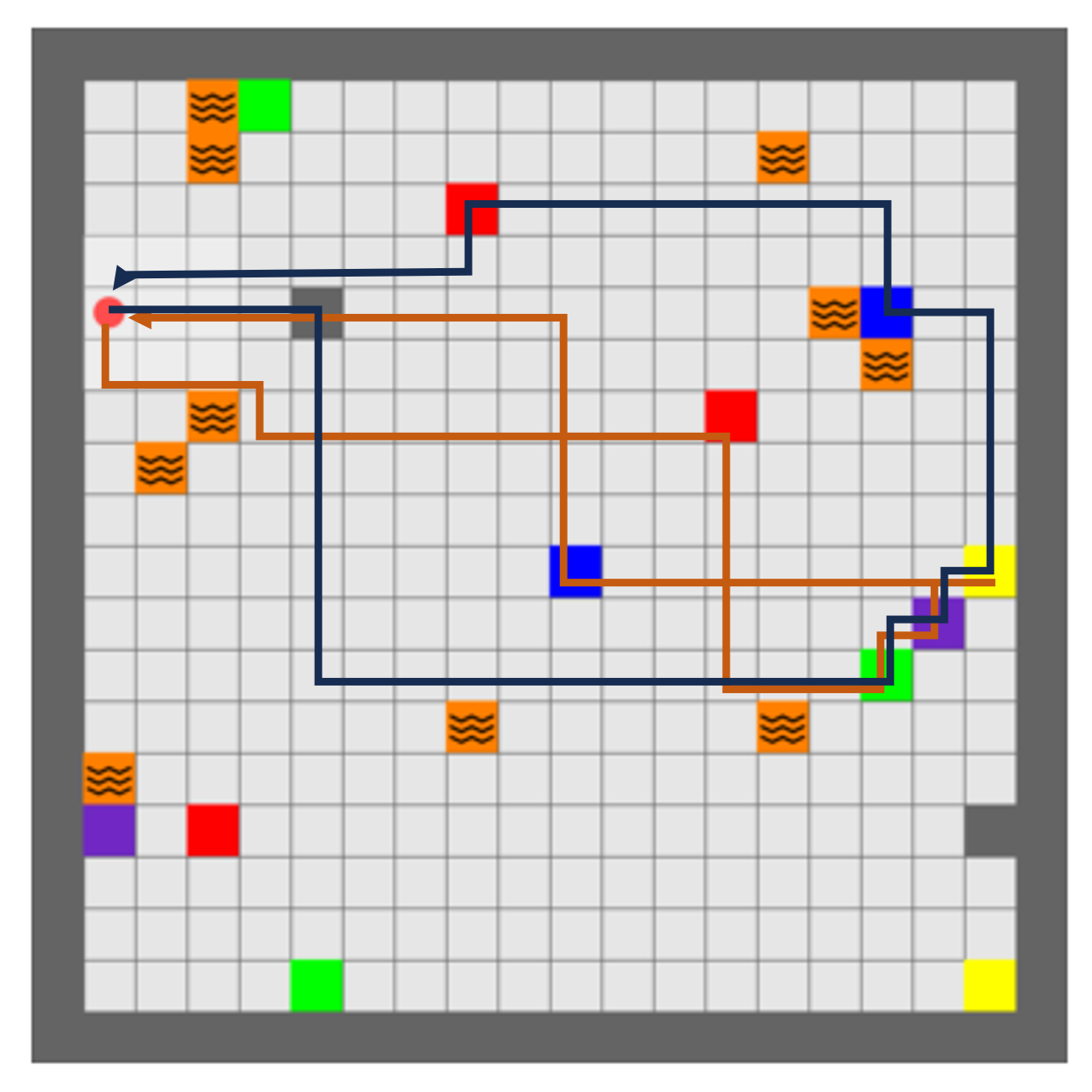}
        \subcaption{Blue/Orange=With/out $\varphi$}
        \label{fig:MinigridExample1}
    \end{minipage}
    \begin{minipage}[t]{0.47\linewidth}
        \centering
        \includegraphics[width=\linewidth]{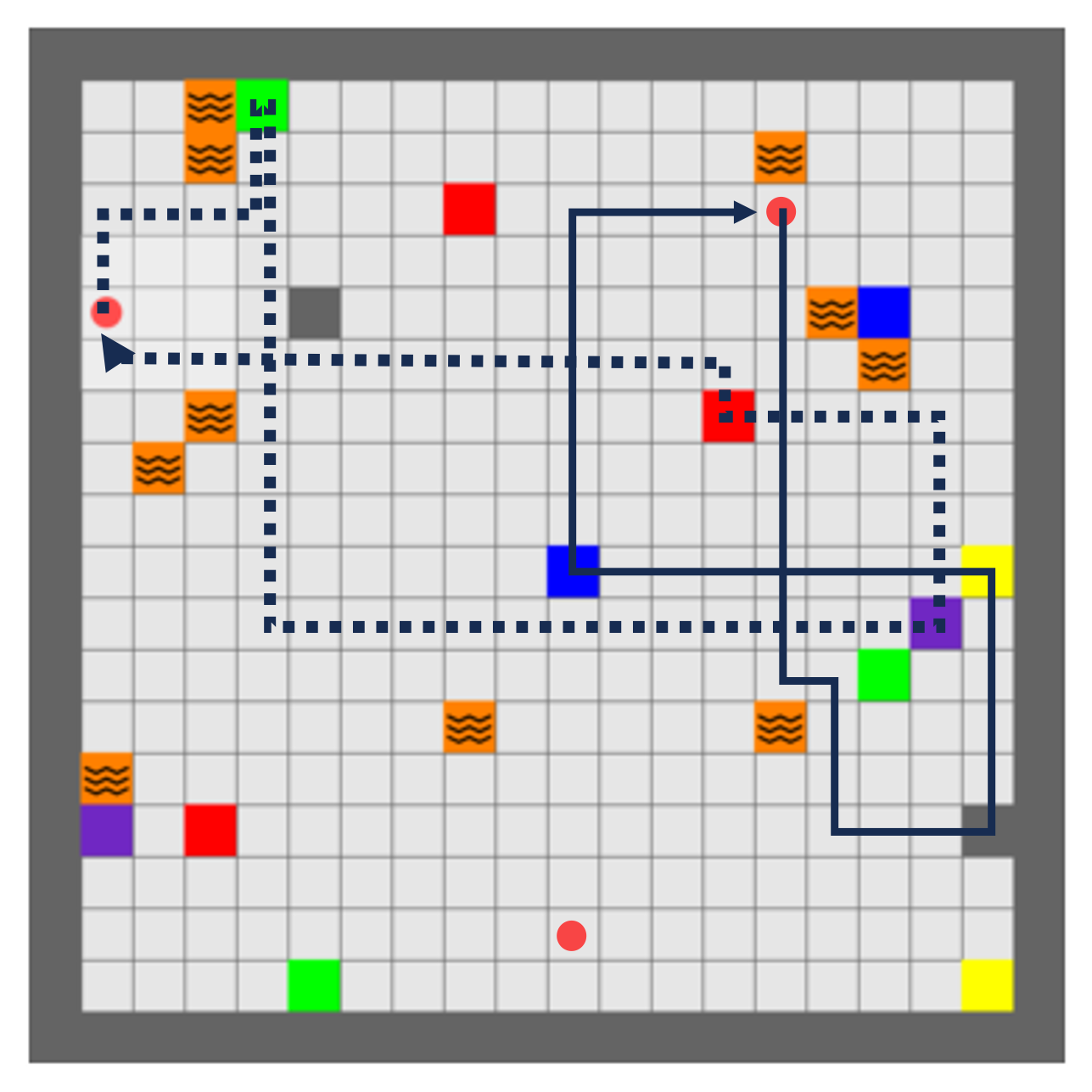}
        \subcaption{Multi-robot trajectories}
        \label{fig:MinigridExample2}
    \end{minipage}
    \caption{Gridworlds with synthesized plans}
    \label{fig:MinigridExperiment}
\end{figure}
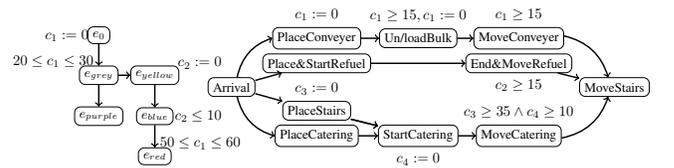
\begin{figure}[b]
    \centering
    \begin{minipage}[t]{0.28\linewidth}
        \centering
        \scalebox{.55}{\begin{tikzpicture}
\matrix[every node/.style={draw=black, rectangle, rounded corners, font=\small}, row sep=15pt, column sep=5pt]{
\node(n1){$e_{0}$};             & \\
\node(n2){$e_{grey}$};          & \node(n4){$e_{yellow}$}; \\
\node(n3){$e_{purple}$};        & \node(n5){$e_{blue}$}; \\
                                & \node(n6){$e_{red}$}; \\
};
\path[->, line width=1pt] (n1) edge (n2)
(n2) edge (n3)
(n2) edge (n4)
(n4) edge (n5)
(n5) edge (n6);
\draw (n1.west)++(-0.5,0.0) node {$c_1 := 0$};
\draw (n2.north)++(-1.1,0.1) node {$20 \leq c_1 \leq 30$};
\draw (n4.east)++(0.5,0.3) node {$c_2 := 0$};
\draw (n5.east)++(0.6,0.0) node {$c_2 \leq 10$};
\draw (n6.north)++(1.1,0.1) node {$50 \leq c_1 \leq 60$};
\end{tikzpicture}}
        \subcaption{TPO$_1$}
        \label{fig:tpo1}
    \end{minipage}
    \hfill
    \begin{minipage}[t]{0.7\linewidth}
        \centering
        \scalebox{.55}{% @reference https://www3.nd.edu/~kogge/courses/cse30151-fa17/Public/other/tikz_tutorial.pdf

\begin{tikzpicture}
% Define and Place nodes
\matrix[every node/.style={draw=black, rectangle, rounded corners, font=\small}, row sep=3pt, column sep=5pt]{
                    & \node(n2){PlaceConveyer}; & \node(n3){Un/loadBulk}; & \node(n4){MoveConveyer}; \\
                    & \node(n5){Place\&StartRefuel}; && \node(n6){End\&MoveRefuel}; &\\
\node(n1){Arrival}; &&&&                                                                \node(n11){MoveStairs}; \\
                    & \node(n8){PlaceStairs}; &  \\
                    & \node(n7){PlaceCatering}; & \node(n9){StartCatering}; & \node(n10){MoveCatering}; \\
};
% Draw Edges
\path[->, line width=1pt] (n1) edge[bend left] (n2)
                          (n2) edge (n3)
                          (n3) edge (n4)
                          (n1) edge (n5)
                          (n5) edge (n6)
                          (n1) edge (n8)
                          (n8) edge (n9)
                          (n8) edge (n9)
                          (n1) edge[bend right] (n7)
                          (n7) edge (n9)
                          (n9) edge (n10)
                          (n4) edge[bend left] (n11)
                          (n6) edge (n11)
                          (n10) edge[bend right] (n11);

% Add fake "nodes" to label the nodes n1,...,n11
\draw (n2.north)++(0,0.3) node {$c_1:=0$};
\draw (n3.north)++(0,0.3) node {$c_1 \geq 15, c_1:=0$};
\draw (n4.north)++(0,0.3) node {$c_1 \geq 15$};
\draw (n6.south)++(0,-0.3) node {$c_2 \geq 15$};
\draw (n8.north)++(0,0.3) node {$c_3:=0$};
\draw (n9.south)++(0,-0.3) node {$c_4:=0$};
\draw (n10.north)++(0,0.3) node {$c_3\geq35 \land c_4\geq10$};
\end{tikzpicture}}
        \subcaption{TPO$_2$}
        \label{fig:tpo2}
    \end{minipage}
    \caption{TPO Specifications for the case studies.}
    \label{fig:TPOs}
\end{figure}

\begin{figure*}[t]
    \centering
    \begin{minipage}[t]{0.24\linewidth}
        \centering
        \includegraphics[width=\linewidth]{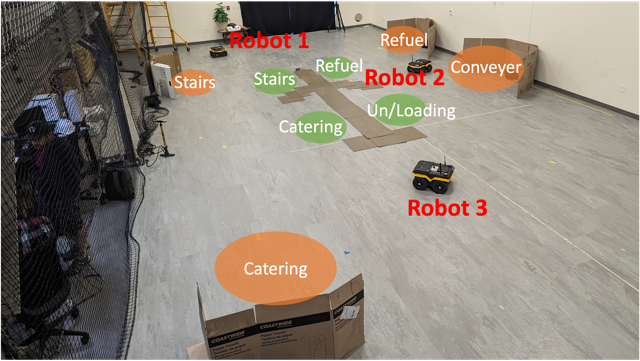}
        \subcaption{Experimental Setup}
        \label{fig:ExperimentalSetup}
    \end{minipage}
    \begin{minipage}[t]{0.24\linewidth}
        \centering
        \includegraphics[width=\linewidth]{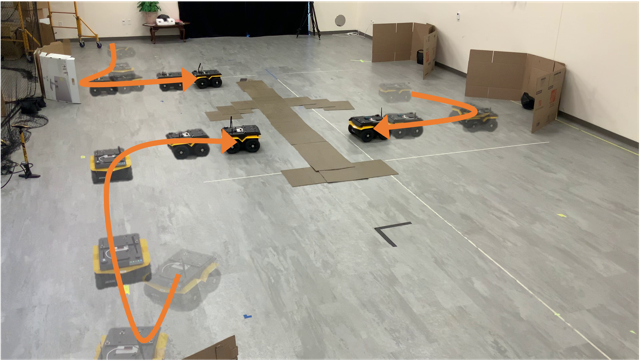}
        \subcaption{Trajectory 1/3}
        \label{fig:Trajectory1}
    \end{minipage}
    \begin{minipage}[t]{0.24\linewidth}
        \centering
        \includegraphics[width=\linewidth]{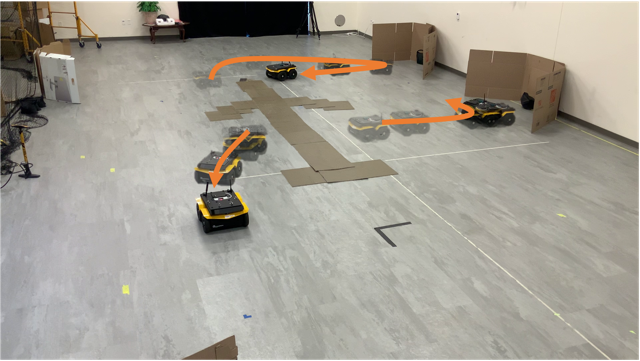}
        \subcaption{Trajectory 2/3}
        \label{fig:Trajectory2}
    \end{minipage}
    \begin{minipage}[t]{0.24\linewidth}
        \centering
        \includegraphics[width=\linewidth]{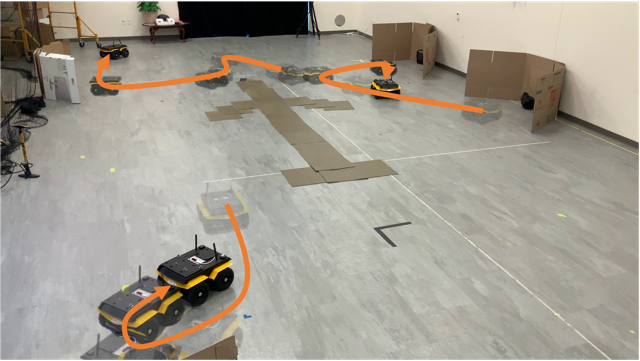}
        \subcaption{Trajectory 3/3}
        \label{fig:Trajectory3}
    \end{minipage}
    \caption{Experimental setup of the aircraft turnaround task (see attached video: \small{\url{https://youtu.be/WUuWFlOoKW8}}).}
    \label{fig:PhysicalExperiment}
\end{figure*}

% \begin{figure}
% \centering
% \scalebox{.8}{\input{sections/06_Experiments/tpo_minigrid}}
% \caption{TPO$_1$ }
% \label{fig:CaseStudyTPO}
% \end{figure}

In this section, we evaluate our algorithm for planning with TPO specification for single and multiple robots on various case studies. 
First, we demonstrate how different timing constraints of TPO cause different robot behaviors. 
Then, we illustrate scalability of the algorithm on a set of benchmarks. 
Lastly, we return to the aircraft turnaround example and show a physical experiment to demonstrate the applicability of the approach to a more realistic scenario.

\subsection{Illustrative Case Studies}

We demonstrate our method on a gridworld environment in \Cref{fig:MinigridExperiment} with multiple colored locations and two TPO tasks: with and without timing constraints.
For the simple task, the robot must visit every color in any order. 
It has the option of visiting any location of the same color but must find a plan with a minimum makespan time.
We ran the algorithm without any constraints and its trajectory is shown in orange in \Cref{fig:MinigridExample1}.
The robot first visits red, green, purple, yellow, blue, and grey in order.
The second task is TPO$_1$ in \Cref{fig:tpo1}.  
% to observe the changes in the behavior.
The robot visited grey, green, purple, yellow, blue and red in order. Observe that the robot now visits the grey first and the red last due to the constraints. 
In \Cref{fig:MinigridExample2}, we extend to two robots the same TPO$_1$ task. The result was two trajectories for the two robots that satisfy TPO$_1$. The trajectories are long, since red has to be visited between $50$ to $60$ time units. 

% % \begin{equation}\label{eq:example-timing-reduced}
% \[\varphi:\ \left(\begin{array}{c}
% t_{g} - t_{0} \in [20, 30] \ \land\ t_{p} - t_{g} \geq 0 \ \land\ t_{y} - t_{g} \geq 0\\
% t_{b} - t_{y} \leq 10 \ \land\ t_{r} - t_{b} \geq 0 \ \land \ t_{r} \in [50, 60] \\
% \end{array}\right)\,\]
% % \end{equation}

\subsection{Benchmarks}
\begin{figure}[b]
    \centering
    \includegraphics[width=\linewidth]{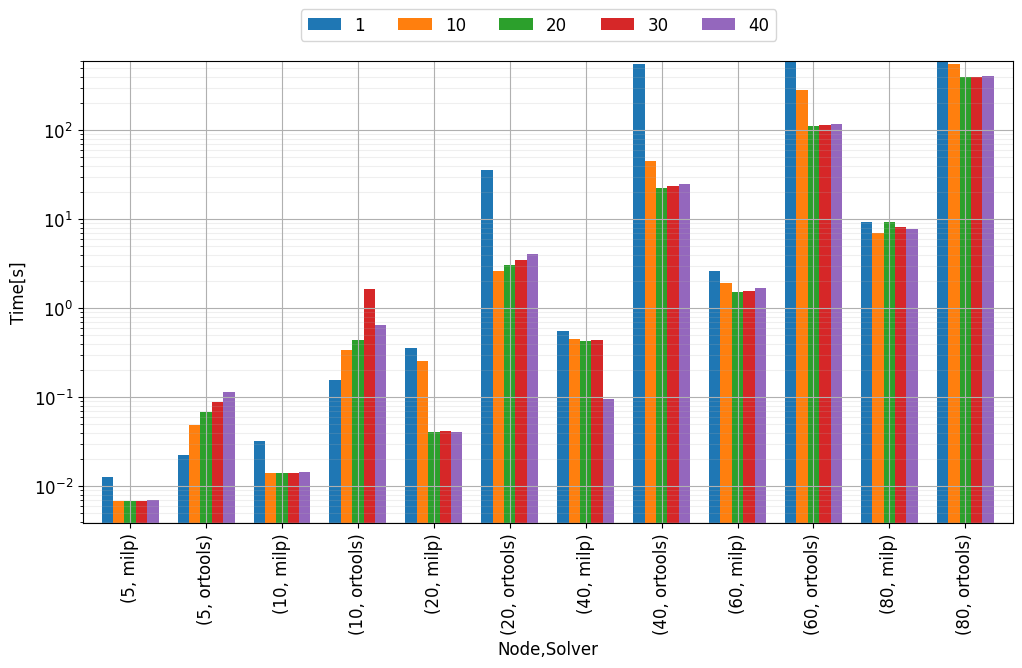}
    \caption{Benchmark results. The x-axis is (\#nodes, solver) and the y-axis is computation time in seconds in log scale. The timeout is set to 600 seconds. The colors of the bars indicate the number of robots (see legend).}
    \label{fig:benchmark}
\end{figure}
Here, we explore how our algorithm scales with the increasing number of states with non-empty labels, timing constraints, and number of robots.
We generated a random set of events varying in number from 5 to 80 in a 30-by-30 gridworld environment.
We incrementally add timing constraints of the form  $d_{i,j}+d_j-\Delta t \leq t_j - t_i \leq d_{i,j}+d_j+\Delta t$ where $\Delta t > 0$ is a constant padding to see how the overall solution time depends on the number of constraints we add and the ``tightness'' of these constraints. 

We ran benchmarks with varying the number of timing constraints (constraints involving $p=25, 50, 75, 100\%$ of all TPO edges),  $\Delta t = 10, 30, 50$, and the number of robots of $1, 10, 20, 30,$ and $40$. We compared our method to heuristic approaches implemented in OR-Tools Routing Libraray \cite{ortools_routing}. The time padding $\Delta t$ and the percentage of the number of edges $p$ did not have significant effects on the results. 
\Cref{fig:benchmark} shows the result at when $p=25\%$ and $\Delta t=30$.

As the number of nodes increases, the problem gets more difficult to solve, taking more time to find the optimal solution. Also, the computation time mostly stays the same as we increase the number of robots. Interestingly, the OR-Tools did not perform well compared to MILP. This is because the heuristics are disabled when unexpected additional constraints (e.g., local timing constraints) are added. Instead, they use constraint programming to solve the problem, which is slower than the Branch and Bound method employed in the MILP solver. 

Also, the algorithm easily scaled up to 80 nonempty-label states as shown in the figure. We further ran the stretch test with a timeout of 30 minutes, and MILP was able to solve up to 160 nodes within $182 \pm 102$ seconds excluding a case (out of 60 runs) when it hit the timeout. It took $210 \pm 232$ seconds including the timeout.
% As the number of timing constraints decreases and the time windows increase (toward top right), the problem gets more difficult to solve, getting closer to a GTSP problem without any constraints. This implies that the constraints help reduce the search space and speed up the MILP computation. Also, the complexity mostly stays the same as we increase the number of robots. Interestingly, the width of the time windows and the number of constraints play a larger role in determining solution time.  Also, the algorithm scales up to 80 nonempty-label states if the timing windows are tight. 

\subsection{Aircraft Turnaround}
We consider the aircraft turnaround example in \Cref{fig:RunningExample}
with robots as ground staff.  The task TPO$_2$ in \Cref{fig:tpo2}, specifies vehicle movement, refuel, and bulk loading/unloading.
%Catering, cleaning, and other in-plane services are assumed to be completed by humans.
% The robots must move vehicles for stairs, bulk loading, catering/cleaning service, and refueling.
%Some events require the ground staff to be present (e.g. bulk loading), 
%and such requirements can be easily embedded by setting the delay to some constant.
The goal is to find the most efficient plans for completing the task according to  TPO$_2$.
In \Cref{fig:PhysicalExperiment}, we show the plans of all robots. 
Robot 1 places the stair truck,  refueling vehicle, and moves out the stair truck.
Robot 2 performs bulk unloading/loading and then moves out the refueling vehicle.
Robot 3 performs the catering services.
Almost all robots finish their assigned events at the same time.  The overall makespan was 59 time units.  We also repeated the same case study for a single robot from various initial states and obtain makespans $152, 155, 163$ time units. %Our approach is effective in utilizing all three available robots. %Hence, we can infer that the tasks were distributed almost equally among the three robots to minimize the makespan. 
A video of this experiment accompanies the submission\footnote{Video: \url{https://youtu.be/WUuWFlOoKW8}}.

% \begin{figure}
% \centering
% \scalebox{.7}{\input{sections/06_Experiments/tpo_aircraft}}
% \caption{TPO$_2$. }
% \label{fig:AircraftTurnaroudTPO}
% \end{figure}

\section{CONCLUSIONS}
We introduce a general framework for planning under Timed Partial Order (TPO) specifications for multiple robots.
Our solution maps the task allocation problem to the Generalized Traveling Salesman Problem (GTSP) with time windows and precedence constraints which we solve by a Mixed-Integer Linear Program (MILP).
Our evaluations of the algorithm on various case studies demonstrate the time-effectiveness of our plans for up to $40$ robots with $160$ nodes.
% and $160$ nodes with $40$ robots.
% \kw{I also ran a stretch test on the MILP approach and the method was able to solve up to $160$ nodes with $40$ robots within 6 minutes. There was 1 case (out of 40 runs at $n=160$) when it hit the timeout of 30 minutes. Should I report this?}

For future work, we plan to investigate variants of the dynamic task assignment problem under TPO specifications and to consider robustness and contingencies in the MILP formulation.  
% Furthermore, we also plan to extend the framework to handle more complex specifications such as Linear Temporal Logic (LTL).

\newpage

\bibliographystyle{plain}
\bibliography{main,lahijanian}

\begin{thebibliography}{10}

\bibitem{alur1994theory}
Rajeev Alur and David~L Dill.
\newblock A theory of timed automata.
\newblock {\em Theoretical computer science}, 126(2):183--235, 1994.

\bibitem{Alur:IEEE:2000}
Rajeev Alur, Thomas~A. Henzinger, Gerardo Lafferriere, George, and J.~Pappas.
\newblock Discrete abstractions of hybrid systems.
\newblock In {\em Proceedings of the IEEE}, volume~88, pages 971--984, 2000.

\bibitem{asarin1998controller}
Eugene Asarin, Oded Maler, Amir Pnueli, and Joseph Sifakis.
\newblock Controller synthesis for timed automata.
\newblock {\em IFAC Proceedings Volumes}, 31(18):447--452, 1998.

\bibitem{chakraa2023optimization}
Hamza Chakraa, Fran{\c{c}}ois Gu{\'e}rin, Edouard Leclercq, and Dimitri
  Lefebvre.
\newblock Optimization techniques for multi-robot task allocation problems:
  Review on the state-of-the-art.
\newblock {\em Robotics and Autonomous Systems}, page 104492, 2023.

\bibitem{Cook+Others/2014/In}
William~J. Cook.
\newblock {\em {In Pursuit of the Traveling Salesman}}.
\newblock Princeton University Press, Princeton, NJ, USA, September 2014.

\bibitem{dechter1991temporal}
Rina Dechter, Itay Meiri, and Judea Pearl.
\newblock Temporal constraint networks.
\newblock {\em Artificial intelligence}, 49(1-3):61--95, 1991.

\bibitem{floyd1962algorithm}
Robert~W Floyd.
\newblock Algorithm 97: shortest path.
\newblock {\em Communications of the ACM}, 5(6):345, 1962.

\bibitem{ortools_routing}
Vincent Furnon and Laurent Perron.
\newblock Or-tools routing library.

\bibitem{Garey+Johnson/1979/Computers}
Michael~R. Garey and David~S. Johnson.
\newblock {\em Computers and Intractability: A guide to the theory of
  {NP}-Completeness}.
\newblock W.H.Freeman, 1979.

\bibitem{gosrich2023multi}
Walker Gosrich, Siddharth Mayya, Saaketh Narayan, Matthew Malencia, Saurav
  Agarwal, and Vijay Kumar.
\newblock Multi-robot coordination and cooperation with task precedence
  relationships.
\newblock In {\em 2023 IEEE International Conference on Robotics and Automation
  (ICRA)}, pages 5800--5806. IEEE, 2023.

\bibitem{guo2023recent}
Huihui Guo, Fan Wu, Yunchuan Qin, Ruihui Li, Keqin Li, and Kenli Li.
\newblock Recent trends in task and motion planning for robotics: A survey.
\newblock {\em ACM Computing Surveys}, 2023.

\bibitem{He:RAL:2019}
Keliang He, Morteza Lahijanian, E~Kavraki, Lydia, and Y~Vardi, Moshe.
\newblock Automated abstraction of manipulation domains for cost-based reactive
  synthesis.
\newblock {\em IEEE Robotics and Automation Letters}, 4(2):285--292, Apr. 2019.

\bibitem{He:ICRA:2015}
Keliang He, Morteza Lahijanian, Lydia~E. Kavraki, and Moshe~Y. Vardi.
\newblock Towards manipulation planning with temporal logic specifications.
\newblock In {\em Int. Conf. Robotics and Automation}, pages 346--352. IEEE,
  May 2015.

\bibitem{Kottinger:AAAI:2024}
Justin Kottinger, Shaull Almagor, Oren Salzman, and Morteza Lahijanian.
\newblock Introducing delays in multi-agent path finding.
\newblock {\em arXiv preprint arXiv:2307.11252}, 2023.

\bibitem{Hadas:ICRA:2007}
H.~Kress-Gazit, G.~Fainekos, and G.~J. Pappas.
\newblock Where's {W}aldo? sensor-based temporal logic motion planning.
\newblock In {\em Int. Conf. on Robotics and Automation}, pages 3116--3121,
  Rome, Italy, 2007. IEEE.

\bibitem{Lahijanian:AR-CRAS:2018}
Hadas Kress-Gazit, Morteza Lahijanian, and Vasumathi Raman.
\newblock Synthesis for robots: Guarantees and feedback for robot behavior.
\newblock {\em Annual Review of Control, Robotics, and Autonomous Systems},
  1:211--236, May 2018.

\bibitem{Lahijanian:ICRA:2009}
Morteza Lahijanian, M.~Kloetzer, S.~Itani, C.~Belta, and S.B. Andersson.
\newblock Automatic deployment of autonomous cars in a robotic urban-like
  environment {(RULE)}.
\newblock In {\em Int. Conf. on Robotics and Automation}, pages 2055--2060,
  Kobe, Japan, 2009. IEEE.

\bibitem{li2016asymptotically}
Yanbo Li, Zakary Littlefield, and Kostas~E Bekris.
\newblock Asymptotically optimal sampling-based kinodynamic planning.
\newblock {\em The International Journal of Robotics Research}, 35(5):528--564,
  2016.

\bibitem{Maler:STL:2004}
Oded Maler and Dejan Nickovic.
\newblock {Monitoring Temporal Properties of Continuous Signals}.
\newblock In {\em Formal Techniques, Modelling and Analysis of Timed and
  Fault-Tolerant Systems: Joint International Conferences on Formal Modeling
  and Analysis of Timed Systmes}, pages 152--166. Springer, 2004.

\bibitem{MalerNP06}
Oded Maler, Dejan Nickovic, and Amir Pnueli.
\newblock From {MITL} to {T}imed {A}utomata.
\newblock In {\em Proceedings of FORMATS}, volume 4202 of {\em LNCS}, pages
  274--289. Springer, 2006.

\bibitem{mingozzi1997dynamic}
Aristide Mingozzi, Lucio Bianco, and Salvatore Ricciardelli.
\newblock Dynamic programming strategies for the traveling salesman problem
  with time window and precedence constraints.
\newblock {\em Operations research}, 45(3):365--377, 1997.

\bibitem{noon1988generalized}
Charles~Edward Noon.
\newblock {\em The generalized traveling salesman problem}.
\newblock University of Michigan, 1988.

\bibitem{nunes2017taxonomy}
Ernesto Nunes, Marie Manner, Hakim Mitiche, and Maria Gini.
\newblock A taxonomy for task allocation problems with temporal and ordering
  constraints.
\newblock {\em Robotics and Autonomous Systems}, 90:55--70, 2017.

\bibitem{parragh2007survey}
Sophie~N Parragh, Karl~F Doerner, and Richard~F Hartl.
\newblock A survey on pickup and delivery problems.
\newblock {\em Part II: Transportation between pickup and delivery locations,
  to appear: Journal f{\"u}r Betriebswirtschaft}, 2007.

\bibitem{pop2023comprehensive}
Petric{\u{a}}~C Pop, Ovidiu Cosma, Cosmin Sabo, and Corina~Pop Sitar.
\newblock A comprehensive survey on the generalized traveling salesman problem.
\newblock {\em European Journal of Operational Research}, 2023.

\bibitem{ren2022conflict}
Zhongqiang Ren, Sivakumar Rathinam, and Howie Choset.
\newblock Conflict-based steiner search for multi-agent combinatorial path
  finding.
\newblock In {\em Proceedings of Robotics: Science and Systems}, 2022.

\bibitem{senderovich2019learning}
Arik Senderovich, Kyle~EC Booth, and J~Christopher Beck.
\newblock Learning scheduling models from event data.
\newblock In {\em Proceedings of the International Conference on Automated
  Planning and Scheduling}, volume~29, pages 401--409, 2019.

\bibitem{sun2022multi}
Dawei Sun, Jingkai Chen, Sayan Mitra, and Chuchu Fan.
\newblock Multi-agent motion planning from signal temporal logic
  specifications.
\newblock {\em IEEE Robotics and Automation Letters}, 7(2):3451--3458, 2022.

\bibitem{suslova2020multi}
Elina Suslova and Pooyan Fazli.
\newblock Multi-robot task allocation with time window and ordering
  constraints.
\newblock In {\em 2020 IEEE/RSJ International Conference on Intelligent Robots
  and Systems (IROS)}, pages 6909--6916. IEEE, 2020.

\bibitem{watanabe2023timed}
Kandai Watanabe, Georgios Fainekos, Bardh Hoxha, Morteza Lahijanian, Danil
  Prokhorov, Sriram Sankaranarayanan, and Tomoya Yamaguchi.
\newblock Timed partial order inference algorithm.
\newblock In {\em Proceedings of the International Conference on Automated
  Planning and Scheduling}, volume~33, pages 639--647, 2023.

\bibitem{williams2013model}
H~Paul Williams.
\newblock {\em Model building in mathematical programming}.
\newblock John Wiley \& Sons, 2013.

\end{thebibliography}

\addtolength{\textheight}{-12cm}   % This command serves to balance the column lengths

%%%%%%%%%%%%%%%%%%%%%%%%%%%%%%%%%%%%%%%%%%%%%%%%%%%%%%%%%%%%%%%%%%%%%%%%%%%%%%%%

%%%%%%%%%%%%%%%%%%%%%%%%%%%%%%%%%%%%%%%%%%%%%%%%%%%%%%%%%%%%%%%%%%%%%%%%%%%%%%%%


% \section*{ACKNOWLEDGMENT}

%%%%%%%%%%%%%%%%%%%%%%%%%%%%%%%%%%%%%%%%%%%%%%%%%%%%%%%%%%%%%%%%%%%%%%%%%%%%%%%%

\end{document}